ORIGINAL ARTICLE

# Warm-started reinforcement learning for iterative 3D/2D liver registration

**Hanyuan Zhang[1] · Lucas He[1] · Zijie Cheng[1] · Abdolrahim Kadkhodamohammadi[2] · Danail Stoyanov[1] · Brian R. Davidson[1] · Evangelos B. Mazomenos[1] · Matthew J. Clarkson[1]**

## Abstract

**Purpose** Registration between preoperative CT and intraoperative laparoscopic video plays a crucial role in augmented reality (AR) guidance for minimally invasive surgery. Learning-based methods have recently achieved registration errors comparable to optimization-based approaches while offering faster inference. However, many supervised methods produce coarse alignments that rely on additional optimization-based refinement, thereby increasing inference time.

**Methods** We present a discrete-action reinforcement learning (RL) framework that formulates CT-to-video registration as a sequential decision-making process. A shared feature encoder, warm-started from a supervised pose estimation network to provide stable geometric features and faster convergence, extracts representations from CT renderings and laparoscopic frames, while an RL policy head learns to choose rigid transformations along six degrees of freedom and to decide when to stop the iteration.

**Results** Experiments on a public laparoscopic dataset demonstrated that our method achieved an average target registration error (TRE) of $15.70 \pm 8.18$ mm, comparable to supervised approaches with optimization, while achieving faster convergence.

**Conclusion** The proposed RL-based formulation enables automated, efficient iterative registration without manually tuned step sizes or stopping criteria. This discrete framework provides a practical foundation for future continuous-action and deformable registration models in surgical AR applications.

## Introduction

In surgery, augmented reality (AR) used alongside intraoperative ultrasound can enhance the visualization of internal liver structures that are not directly visible during the procedure, such as tumor locations and vascular distributions [1, 2]. While ultrasound remains the standard intraoperative imaging modality, AR provides complementary visual cues by overlaying preoperative CT information onto the laparoscopic view, thereby improving spatial awareness. This process requires registration techniques to align preoperative CT volumes with intraoperative laparoscopic video images [3].

✉ Hanyuan Zhang
hanyuan.zhang.23@ucl.ac.uk

1 UCL Hawkes Institute, University College London, London, UK

2 Medtronic plc., London, UK

However, liver deformation and the limited visibility of the liver surface introduce significant ambiguity during the registration process. In particular, global registration methods are prone to converge to incorrect local minima due to false correspondences. In such cases, rigid initialization plays a crucial role by constraining the search space to a region close to the true alignment and increasing the likelihood of converging to the correct global minimum, thereby improving both robustness and accuracy.

## Background

Early approaches to CT-to-video registration primarily relied on 3D–3D registration, which involved reconstructing intraoperative liver surfaces using laser range scanners [4], intraoperative CT [5], or stereo laparoscopes [6, 7]. However, these methods are difficult to deploy in clinical practice: laser scanners are incompatible with trocar ports, intraoperative CT is seldom available, and stereo laparoscopes, although enabling techniques such as globally optimal ICP (Go-ICP) [7] and anterior ridge matching [6], require stereo laparoscopes, whereas standard laparoscopic setups rely on monocular cameras. Consequently, recent research has focused on methods compatible with monocular laparoscopes, although accurate depth recovery from single-view images remains challenging [8, 9].

More recent work has focused on 3D–2D registration, aligning 2D intraoperative liver contours with projected 3D models. Liver features such as ridges, ligaments, and silhouettes are commonly used [6], and PnP-based initialization combined with RANSAC provides robust rigid alignment [10]. Learning-based methods, such as simulation-based registration frameworks [11], aim to bridge the gap between 3D anatomical models and 2D intraoperative observations but often encounter domain shifts between synthetic and real data. Other approaches, including deep-hashing pose estimation [12] and patient-specific pre-trained alignment schemes [13], improve automation and inference speed but still face challenges such as discrete pose representation, limited validation on patient data, and sensitivity to occlusions.

While recent studies have also explored end-to-end non-rigid registration frameworks [14–16], rigid alignment remains a critical prerequisite for achieving reliable initialization, especially under large pose deviations and partial visibility. Overall, contour-based registration remains limited by sparse geometric constraints and inherent depth ambiguities under partial visibility.

A recent study further extended feature-based registration by introducing an additional depth channel generated using DepthAnything V2, which was fused with contour information within the network [17]. During inference, this method adopts a multi-iteration strategy to progressively refine the registration results, yielding better accuracy than contour-based features alone. However, repeated iterations introduce substantial computational overhead due to frequent data transfers between the CPU and GPU, and its termination criterion remains relatively simple.

To address these limitations, we propose a reinforcement learning (RL)-based registration framework. We design an interactive environment in which, given the current camera pose, new contours and depth maps are rendered on the fly. An RL agent then determines how to adjust the pose based on the current and target observations. We begin by validating this framework using a discrete-action space, demonstrating the feasibility of learning-based control for iterative CT-to-video registration.

Our main contributions are summarized as follows:

1. We design a discrete-action reinforcement learning environment for camera pose adjustment in rigid registration, implemented with GPU acceleration for efficient training and inference.
2. The proposed RL agent automatically selects motion direction, step size, and termination condition, enabling fully automatic iterative CT-to-video registration *given pre-segmented anatomical masks/contours*, with improved convergence efficiency.

## Method

We developed a reinforcement learning (RL) network to solve the registration task. The agent's input includes two sets of observations: the target (intraoperative) contour, mask, and depth images, and the current corresponding images, which are rendered from the current estimated camera pose. At the start of each episode, this pose is randomly sampled to ensure robust initialization. The agent outputs a discrete-action specifying four components: the 6 DoF parameter to adjust, the direction (positive or negative), the step magnitude (large or small), and whether to terminate the episode. Upon receiving the action command, the environment updates the camera pose accordingly and renders a new set of contour, mask, and depth images. These new renderings serve as the current state for the agent's next decision, and the process continues iteratively until the agent issues a termination command or a maximum number of steps is reached. The complete pipeline is illustrated in Fig. 1.

### Environment design

*Geometry and rendering pipeline*: A task-specific rendering environment is constructed to provide reliable geometric supervision while maintaining computational efficiency. Anatomical surfaces, including the liver silhouette, left and right ridges, and ligament are imported from `VTK` files and merged into a single `PyTorch3D` mesh. A face-to-component mapping is preserved to recover per-structure binary masks from a single rasterization pass.

A Laparoscope model is defined using intraoperative intrinsic parameters $(f_x, f_y, c_x, c_y)$, downsampled to the training resolution $(H = W = 128)$ by linearly scaling the intrinsics according to the resolution ratio. Depth and `pix_to_face` buffers are cached for reuse across frames. RGB shading is disabled by default, as the network primarily relies on shape and depth cues for pose estimation.

*Observation space*: During the training stage, at each step, the environment provides a paired observation $\left(\mathbf{I}^{\mathrm{t}}, \mathbf{I}^{\mathrm{tgt}}\right)$, where both are $6 \times H \times W$ tensors rendered from the liver mesh. The six channels include:

1. Channels 1–4: semantic masks of the ligament, right ridge, left ridge, and liver silhouette;
2. Channel 5: normalized inverse depth within the liver mask.

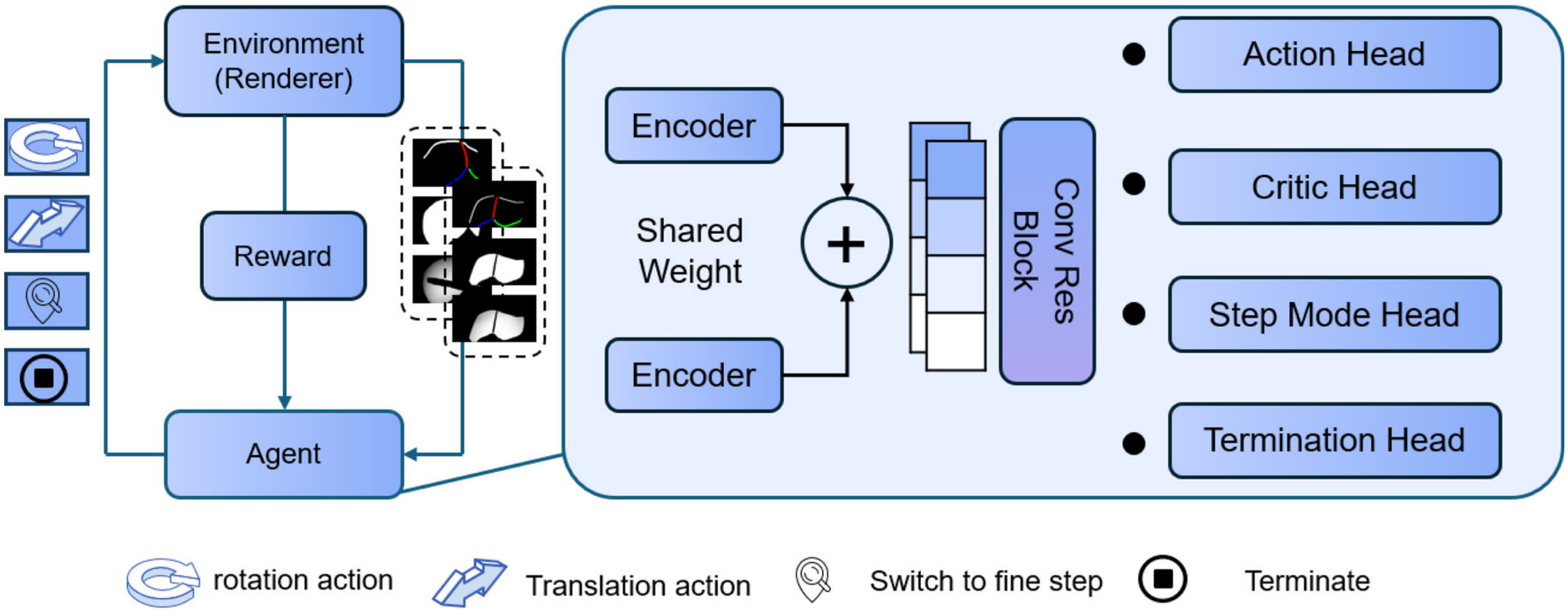

**Fig. 1** Overview of the proposed framework. The environment generates a rendered source image, which together with the intraoperative target forms two six-channel inputs to the agent. The agent outputs an action, step granularity, and termination probability. The environment updates the pose and assigns rewards based on MSE change. The updated observation is fed back iteratively. The network adopts a Siamese architecture with shared-weight encoding, residual convolutional blocks, and four output heads

3. Channel 6: binary liver mask.

We emphasize that Channel 5 is not used as a metrically accurate reconstruction of the liver surface. Instead, it serves as a relative geometric cue that complements the semantic contour/mask channels by providing depth ordering and coarse shape information within the visible liver region. To reduce sensitivity to unknown global scale/bias (e.g., when depth is estimated by a monocular network), we normalize depth within the liver mask for each frame: we first convert to inverse depth and then apply per-frame min–max normalization (with optional percentile clipping) inside the liver mask, setting pixels outside the mask to zero. With this design, the agent primarily relies on relative depth variations and discontinuities aligned with anatomical boundaries, rather than absolute metric depth values.

The target view $\mathbf{I}^{\mathrm{tgt}}$ is rendered once per episode from a randomly sampled viewpoint ensuring sufficient surface visibility, and is cached for efficiency. The current observation $\mathbf{I}^{\mathrm{t}}$ is re-rendered at each step according to the updated camera pose. To improve generalization, random augmentations are applied on the simulated target image: Morphological operations (dilation/erosion) and random erasing on channels 1–4 are applied to simulate occlusions caused by surgical instruments. synthetic occlusion and depth normalization on channel 5, and additional morphological operations on channel 6. An optional binary mask may be applied to emulate the circular field of view of a laparoscope and to suppress irrelevant background regions outside the visible area.

*Action space and SE(3) transition*: The agent operates in a 12-dimensional discrete-action space:

$$\mathcal{A} = \{\pm x,\ \pm y,\ \pm z,\ \pm r_x,\ \pm r_y,\ \pm r_z\}.$$

The SE(3) exponential formulation unifies translation and rotation within a single, manifold-consistent representation. Unlike quaternion or axis–angle forms, it operates directly in the Lie algebra tangent space, ensuring linear, stable, and symmetry-preserving updates [18]. This representation naturally integrates with SE(3)-equivariant learning and geometric control frameworks, enabling efficient and robust policy optimization [19].

$$\mathbf{T}(\boldsymbol{\xi}) = \begin{bmatrix} \exp([\boldsymbol{\omega}]_\times) & \mathbf{V}\mathbf{v} \\ \mathbf{0}^\top & 1 \end{bmatrix}, \quad \mathbf{V} = \mathbf{I} + \frac{1-\cos\theta}{\theta^2}[\boldsymbol{\omega}]_\times + \frac{\theta - \sin\theta}{\theta^3}[\boldsymbol{\omega}]_\times^2, \tag{1}$$

where $\theta = \|\boldsymbol{\omega}\|$ and $[\cdot]_\times$ is the skew-symmetric operator. The rotation matrix is orthonormalized via singular value decomposition (SVD) to ensure $\det(\mathbf{R}) = 1$. Adaptive step sizes are supported: coarse steps handle large misalignments, while fine steps refine near convergence. The step type can also be predicted by a dedicated policy head.

*Pose error and reward function:* Pose accuracy during training is quantified by the mean squared error (MSE) of model points transformed by the relative pose between the current and target views. The relative transformation $\mathbf{T}_{\mathrm{rel}} \in \mathrm{SE}(3)$ is obtained as

$$\mathbf{T}_{\mathrm{rel}} = \mathbf{T}_{\mathrm{tgt}}^{-1}\mathbf{T}_t = \begin{bmatrix} \mathbf{R}_{\mathrm{rel}} & \mathbf{t}_{\mathrm{rel}} \\ \mathbf{0}^\top & 1 \end{bmatrix},$$

The pose error at time step $t$ is then computed as:

$$m_t = \frac{1}{|\mathcal{P}|} \sum_{p_i \in \mathcal{P}} \|\mathbf{p}_i - (\mathbf{R}_{\text{rel}}\mathbf{p}_i + \mathbf{t}_{\text{rel}})\|_2^2 . \tag{2}$$

where $\mathcal{P} = \{\mathbf{p}_i\}$ denotes 3D points (mesh vertices) from the *preoperative* CT-derived liver surface model, expressed in the liver model coordinate frame.

This MSE serves as a proxy for rigid-pose discrepancy and is only used as the internal reward signal during training. In real-data evaluation, we report TRE using the official protocol of Rabbani et al., computed between the LUS-derived tumor profile (ground truth) and the CT tumor volume projected by the estimated pose. These tumor annotations are used only for evaluation and are never provided to the model during inference.

We define a step reward with a final bonus that encourages monotonic convergence. We use a slightly stronger penalty for non-improving actions ($-1.1$ vs. $+1$) to discourage dithering and encourage efficient trajectories with fewer corrective steps (consistent with our goal of reducing inference iterations).

At each step, improvement is measured as $\Delta m_t = m_{t-1} - m_t$, and the reward is

$$r_t^{\text{step}} = \begin{cases} +1, & \text{if } \Delta m_t > 0, \\ -1.1, & \text{if } \Delta m_t < 0, \\ 0, & \text{otherwise.} \end{cases} \tag{3}$$

Episodes terminate when $m_t < m_{\text{succ}}$, when the view becomes empty, or upon reaching the step limit. A final bonus is applied at termination to encourage global improvement:

$$r^{\text{final}} = \beta \cdot \max\left(0, \frac{m_0 - m_T}{m_0 + \varepsilon}\right), \tag{4}$$

where $m_0$ is the initial error, $\beta$ is a scaling factor, and $\varepsilon$ prevents division by zero.

## Network architecture

The agent follows an actor–critic design with a shared Siamese encoder. At each step, the current rendered observation $\mathbf{I}^t$ and the target observation $\mathbf{I}^{\text{tgt}}$ are processed by a shared-weight RefineNet backbone pretrained on a supervised pose regression task [17]. The resulting feature maps are fused channel-wise (concatenation), followed by a residual convolutional block and global average pooling to obtain a 512-dimensional embedding. A lightweight MLP projects the embedding into four heads: (i) an actor producing logits over 12 discrete SE(3) actions, (ii) a critic estimating the state value, (iii) a binary head predicting step granularity (coarse vs. fine), and (iv) a binary head predicting termination probability.

## Training

*Warm-start and optimization.* We train the policy using Proximal Policy Optimization (PPO). The RefineNet encoder is warm-started from supervised pretraining [17] and kept frozen during RL to stabilize feature extraction, prevent catastrophic forgetting, and accelerate convergence. Only the policy/critic heads and the subsequent fusion layers are optimized.

Each PPO update uses rollouts of 4096 environment steps, followed by four optimization epochs with mini-batches of size 128. We use Adam with $\text{lr} = 10^{-4}$, discount factor $\gamma = 0.98$, GAE parameter $\lambda = 0.95$, and clipping parameter $\epsilon = 0.3$. Gradient norms are clipped to 0.5 for stability. Entropy regularization is linearly annealed from 0.05 to 0.00 to balance exploration and exploitation.

*Loss function.* The overall objective consists of the clipped surrogate loss, value regression, entropy bonus, and auxiliary supervision:

$$\mathcal{L} = \mathcal{L}_{\text{PPO}} + \lambda_v \mathcal{L}_{\text{value}} - \lambda_e \mathcal{H} + \lambda_d (\mathcal{L}_{\text{step}} + \mathcal{L}_{\text{term}}),$$

where $\mathcal{L}_{\text{step}}$ and $\mathcal{L}_{\text{term}}$ are binary cross-entropy losses for the step-mode and termination heads.

*Curriculum on the success threshold.* To ease learning under large initial misalignment, we adopt a curriculum by progressively tightening the success threshold $m_{\text{succ}}$ (measured in $\text{mm}^2$ as in Eq. 2). We start from a loose threshold and reduce it stage-wise to encourage increasingly accurate alignment. In our implementation, $m_{\text{succ}}$ is reduced from $500 \to 300 \to 100 \to 10\ \text{mm}^2$, where $\sqrt{m_t}$ corresponds to the average point-wise RMSE in mm (e.g., $100\ \text{mm}^2 \to 10$ mm).

*Auxiliary heads supervision.* The step-mode head is supervised using the current pose error $m_t$: we label *coarse* when $m_t > 200\ \text{mm}^2$ and *fine* otherwise. The termination head uses a stricter margin: frames with $m_t < 50\ \text{mm}^2$ are labeled as positive, and all others as negative. These labels are derived from synthetic training signals only and are never required at test time on real data.

*Episode Termination and Invalid Actions.* Episodes terminate upon success ($m_t < m_{\text{succ}}$), reaching the step limit (256 steps), entering an empty-view state, or when the policy predicts termination. Actions leading to an empty-view trigger an "undo" mechanism that reverts invalid motions.

### Inference

During inference, both the coarse/fine step head and the termination head are activated to guide the registration process autonomously. A random initial camera pose with visible liver surface is first selected as the starting point. The policy then iteratively updates the camera pose according to the predicted SE(3) action and the selected step mode (coarse or fine). At each iteration, the termination head outputs a probability indicating whether the current alignment should stop. The process automatically terminates once this probability exceeds a predefined threshold, indicating convergence. If the threshold is not reached within the maximum step limit, the final registration result is chosen as the frame with the highest termination confidence among all iterations.

## Experiment

s by Rabbani et al. [20] (also used in subsequent studies such as Mhiri et al. [21]), which contains intraoperative ultrasound and laparoscopic imaging data from four patients, with 8–21 frames per case. Tumor locations were identified intraoperatively using ultrasound, providing a quantitative ground truth for validating AR-based guidance via tumor TRE.

Consistent with previous reports [15, 16], *Patient 2* is widely regarded as particularly challenging for rigid CT-to-video registration. Prior analyses attribute this to a substantial torsion/deformation during image acquisition, which alters the liver geometry and violates the rigid-model assumption. As a result, even expert manual alignment yields a high average error (approximately 35 mm), and optimization-based pipelines typically fail to reach clinically acceptable precision. Because many learning-based methods are trained on synthetic renderings generated from such optimization-driven rigid pipelines, their attainable performance on this case is inherently limited.

Importantly, in this work we *do not exclude Patient 2*. Instead, we report results both *including* and *excluding* *Patient 2*, following common practice [14–16, 21], and we additionally provide complementary analyses to help interpret performance under this known outlier scenario. Our primary discussion focuses on *Patients 1, 3, and 4*, while *Patient 2* is included as a stress test highlighting the limitations of rigid alignment and tumor-only evaluation metrics.

*Training cases and synthetic pose generation.* Following prior work [14–16], we train patient-specific models on 4 cases Starting from a reference pose with all anatomical features visible, we generate 500,000 candidate camera poses per patient by sampling rotations in $[-20°, 20°]$ and translations in $[-50, 50]$ mm along each axis. We retain only poses that render at least two distinct anatomical structures to ensure informative supervision.

*Episode construction.* For each training episode, we randomly sample two retained poses: one defines the target observation $\mathbf{I}^{\text{tgt}}$ (cached for efficiency), and the other defines the initial current observation $\mathbf{I}^{t}$. The agent iteratively updates the pose to align $\mathbf{I}^{t}$ to $\mathbf{I}^{\text{tgt}}$ in the simulated environment.

*Training schedule.* We adopt the curriculum described in Sect. Training by progressively tightening the success threshold $m_{\text{succ}}$. After the policy becomes stable, we enable the step-mode and termination heads and train them using the synthetic supervision signals defined in Sect. Training. Unless otherwise stated, we use the PPO hyper-parameters in Sect. Training.

*Hardware and implementation.* All models are implemented in PyTorch and trained on NVIDIA RTX 4090 GPUs.

### Synthetic data evaluation

After training, we additionally generated 1000 pose pairs per patient (for the four patients used in training) that were completely unseen during training. During validation, both the *step-mode* and *termination* heads were activated. We varied the termination output threshold between 0.5 and 0.9 in increments of 0.1 to examine whether a stricter confidence requirement could lead to lower registration error.

### Real-data evaluation

In the real-data experiments, we used manually segmented data from 4 patients as inputs. Considering the trade-off between computation time and registration accuracy, the termination confidence threshold was empirically set to 0.8.

For each frame, ten random initial poses were selected for registration, and the target registration error (TRE) was computed for quantitative evaluation. The initial poses were distributed around either the center or the edges of the image, covering both the entire liver and partial liver regions.

All experiments were implemented in PyTorch and executed on NVIDIA RTX 4090 GPUs. The code and pretrained models will be made publicly available upon acceptance.

## Results

### Synthetic evaluation result

As shown in Fig. 2, tightening the termination confidence threshold consistently reduces the registration error across all four synthetic patient datasets. For *Patient 1*, the same downward trend is observed; however, its overall MSE remains

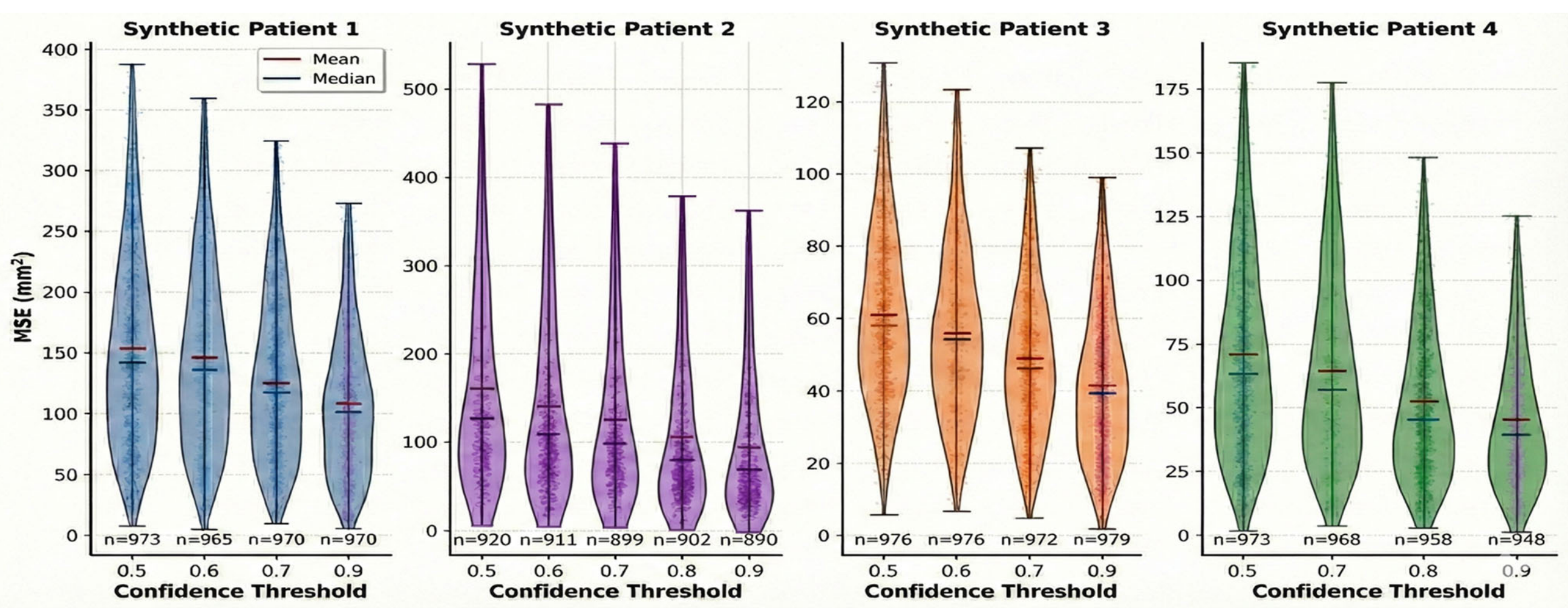

**Fig. 2** Violin plots showing the distribution of mean squared errors (MSE) across different confidence thresholds (0.5–0.9) for four synthetic patients. Each violin depicts the MSE distribution after outlier removal, with the red line indicating the mean and the blue line indicating the median

substantially higher than those of *Patient 3* and *Patient 4*. During training, we also observed that the number of successful episodes for *Patient 1* and *Patient 2* were significantly lower, suggesting a more challenging optimization landscape. In contrast, *Patients 3* and *4* exhibit stable and expected behavior: although the training process stops automatically when the MSE reaches 10 $mm^2$, the termination head was designed to output `true` when the predicted MSE falls below 50 $mm^2$. With a confidence threshold of 0.8, more than half of the samples for *Patients 3* and *4* achieve MSE values below 50 $mm^2$ (median), indicating reliable convergence. However, increasing the threshold also prolongs inference, from approximately 10 min to over 30 min for processing 1,000 registration pairs (≈0.6–1.8 s per registration) on an RTX 4070 Laptop.

Overall, stricter thresholds lead to lower registration errors but at the cost of increased computational time.

## Real data evaluation result

As shown in Table 1, the proposed discrete RL-based method outperforms the compared approaches on *Patient 2*, *Patient 3* and *Patient 4*. Across ten repeated experiments, the standard deviations of the TRE remain small for these three cases, indicating stable performance. However, for *Patient 1*, the results are less consistent: although the mean TRE is moderate, its variability is noticeably higher, reflecting the instability observed during training, where *Patient 1* also showed inferior convergence compared to *Patients 3* and *4*.

Notably, this observation is consistent with the synthetic data evaluation, suggesting a positive correlation between synthetic and real-world performance. Furthermore, when the model was trained without pretrained feature extractor weights, it failed to converge even after 24 h of training, showing no sign of stabilization. In contrast, a warm-started initialization could led to faster convergence within approximately 10 h.

During inference, our RL agent typically terminates within ∼40–256 refinement steps, corresponding to ∼1–5 s on an RTX 4070 Laptop under our implementation settings. This runtime is substantially lower than we observe for optimization-based refinement baselines in the same setup (e.g., ∼30–60 s for our CMA-ES Hausdorff optimization [17]).

For reference, Labrunie et al. [16] report that their fully automatic rigid initialization (pose estimation) takes 10–40 s. Hybrid pipelines that combine a predictor with a geometric/differentiable refinement stage can be faster: ADeLiR reports an average runtime of 4.90 s in total (0.05 s coarse affine + 4.85 s fine RTO with ∼10 iterations) on an RTX 2080 Ti [14].

In addition, our ablation study evaluates inference under two settings, with and without step-size mode, and reports the corresponding registration errors. Since the segmentations are manually produced, we further perturb the manual masks by randomly removing true positives and adding false positives to simulate errors that may arise from real automatic segmentation. Both sets of results are summarized in Table 2. As shown, enabling step-size mode yields lower registration error than disabling it, and using noise-perturbed segmentation masks achieves errors comparable to the noise-free setting.

## Discussion and conclusion

Our results demonstrate the feasibility of using reinforcement learning (RL) networks with discrete-action spaces for reg-

**Table 1** Target tumor registration error (TRE, mm) across patients

| Method | P1 | P2 | P3 | P4 | Avg | Avg w/o P2 |
|---|---|---|---|---|---|---|
| MA | 15.14 | 35.48 | 30.48 | 16.29 | 24.35 | 20.63 |
| LMR [15] | 17.40 | 53.80 | 17.6 | 17.0 | 26.45 | 17.33 |
| NM [21] | 14.82 | 51.43 | 20.15 | 12.95 | 24.83 | 15.87 |
| Opt-B [16] | 14.87 | N/A | 22.40 | 7.23 | – | 14.83 |
| ADeLiR [14] | **12.45** | 46.44 | 17.13 | 11.20 | 21.81 | 13.60 |
| Discrete RL | 16.60 ± 5.40 | **27.36** ± 2.19 | **11.94** ± 1.20 | **6.90** ± 0.77 | **15.70** ± 8.18 | **11.81** ± 5.09 |

Bold values indicate the lowest mean TRE value for each patient or average column
MA: manual initialization. Averages are given for all images available in the dataset of each patient

**Table 2** Registration error statistics (mm)

| | P1 w/o st | P1 | P1 noise | P2 w/o st | P2 | P2 noise |
|---|---|---|---|---|---|---|
| **Mean** | 17.5979 | **16.6012** | 20.4837 | 28.8995 | 27.3645 | **25.4129** |
| **Max** | **24.2514** | 24.3671 | 26.2055 | 100.4330 | 95.4733 | **84.0417** |

| | P3 w/o st | P3 | P3 noise | P4 w/o st | P4 | P4 noise |
|---|---|---|---|---|---|---|
| **Mean** | 15.3172 | **11.9413** | 15.1642 | 10.2907 | **6.8980** | 6.9092 |
| **Max** | 21.6206 | **14.0890** | 22.4790 | 13.2877 | 8.8481 | **8.4765** |

For each patient, three columns are reported: w/o st (no step-size switching), with step-size switching, and noise (TRE after randomly removing some true positives and injecting false positives into the manual segmentation). Top: Patients 1 & 2; Bottom: Patients 3 & 4. The lowest value is in bold

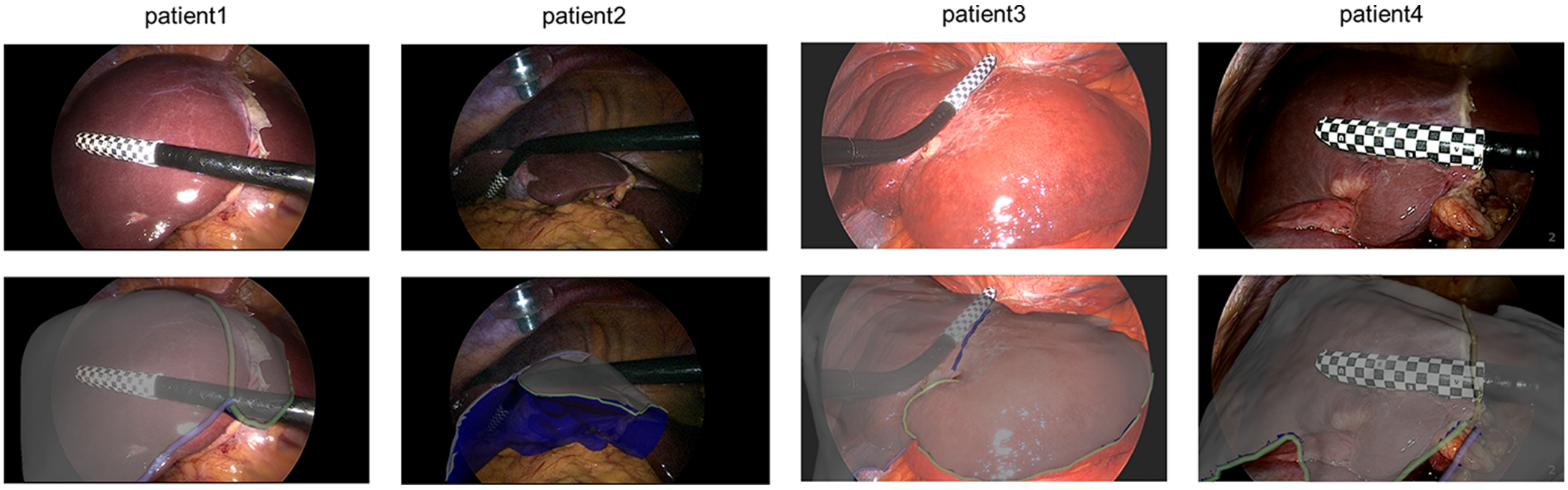

**Fig. 3** Example of registration results: from left to right are patient 1, 2, 3, and 4. The top row shows the intraoperative videos, and the bottom row shows the images with the overlaid 3D meshes

istration tasks. Compared with conventional optimization-based methods, our approach reduces computational overhead. During inference on real data, most cases converge to the target confidence within 40 steps. However, performance is currently constrained by the discrete step sizes. In future work, we plan to extend the method to continuous actions. This would enable adaptive step-sizing, larger steps in the initial stages to accelerate convergence and finer steps near the optimum potentially reducing inference time while improving registration accuracy. However, this transition introduces optimization challenges, as the policy must be trained over a continuous action space.

At the current stage, our model remains patient-specific. This design is advantageous for rapid training when only a few patient datasets are available. However, as the number of patients grows, training separate models becomes inefficient. A more scalable approach would be to develop a foundation model by registering multiple liver datasets into a common coordinate system, which would be more effective for capturing shared anatomical structures and enabling faster adaptation, or potentially removing the need for fine-tuning altogether.

We also note that, although Patient 2 shows a clear reduction in tumor TRE compared to other methods, the qualitative (visual) alignment is not as satisfactory as expected. This dis-crepancy arises because our reported error is computed only on tumor landmarks; thus, a case can achieve a low tumor TRE even when the overall liver surface alignment remains imperfect (Fig. 3).

In addition, the positive correlation between synthetic and real data performance suggests that our current data augmentation strategies partially mitigate the domain gap. Nevertheless, the synthetic evaluation converges faster, requiring fewer iterations to reach comparable performance. This observation indicates that synthetic domains remain less complex than real-world data, and domain adaptation techniques could therefore play a crucial role in further narrowing the domain gap [14].

In conclusion, we demonstrated the feasibility of automatic iterative registration using a foundational discrete action–based reinforcement learning framework. Future work will focus on extending this framework to a continuous action space for more precise motion control and on further reducing the domain gap between synthetic and real data, ultimately aiming to develop a patient-generic model.

**Acknowledgements** This work is supported in part by the National Institute for Health Research (NIHR) under its Invention for Innovation (i4i) Programme (Grant Reference Number NIHR II-LA-1116-20005). The views expressed are those of the author(s) and not necessarily those of the NIHR or the Department of Health and Social Care. This work is additionally supported by the Wellcome/EPSRC Centre for Interventional and Surgical Sciences (WEISS) [203145Z/16/Z], and EPSRC under the "Human-centred Machine Intelligence to optimise Robotic Surgical Training (HuMIRoS)" project [EP/Z534754/1] and the EPSRC [EP/T029404/1]. HZ would also like to thank Medtronic for supporting his PhD studentship.

## Declarations

**Conflict of interest** The authors declare that they have no conflict of interest.

## References

1. Ramalhinho J, Yoo S, Dowrick T, Koo B, Somasundaram M, Gurusamy K, Hawkes DJ, Davidson B, Blandford A, Clarkson MJ (2023) The value of augmented reality in surgery—a usability study on laparoscopic liver surgery. Med Image Anal 90:102943
2. Schneider C, Thompson S, Totz J, Song Y, Allam M, Sodergren M, Desjardins A, Barratt D, Ourselin S, Gurusamy K et al (2020) Comparison of manual and semi-automatic registration in augmented reality image-guided liver surgery: a clinical feasibility study. Surg Endosc 34(10):4702–4711
3. Ali S, Espinel Y, Jin Y, Liu P, Güttner B, Zhang X, Zhang L, Dowrick T, Clarkson MJ, Xiao S et al (2025) An objective comparison of methods for augmented reality in laparoscopic liver resection by preoperative-to-intraoperative image fusion from the miccai2022 challenge. Med Image Anal 99:103371
4. Fusaglia M, Hess H, Schwalbe M, Peterhans M, Tinguely P, Weber S, Lu H (2016) A clinically applicable laser-based image-guided system for laparoscopic liver procedures. Int J Comput Assist Radiol Surg 11(8):1499–1513
5. Pelanis E, Teatini A, Eigl B, Regensburger A, Alzaga A, Kumar RP, Rudolph T, Aghayan DL, Riediger C, Kvarnström N et al (2021) Evaluation of a novel navigation platform for laparoscopic liver surgery with organ deformation compensation using injected fiducials. Med Image Anal 69:101946
6. Robu MR, Ramalhinho J, Thompson S, Gurusamy K, Davidson B, Hawkes D, Stoyanov D, Clarkson MJ (2018) Global rigid registration of CT to video in laparoscopic liver surgery. Int J Comput Assist Radiol Surg 13(6):947–956
7. Luo H, Yin D, Zhang S, Xiao D, He B, Meng F, Zhang Y, Cai W, He S, Zhang W et al (2020) Augmented reality navigation for liver resection with a stereoscopic laparoscope. Comput Methods Programs Biomed 187:105099
8. Zhang Y, Zou Y, Liu PX (2024) Point cloud registration in laparoscopic liver surgery using keypoint correspondence registration network. IEEE Trans Med Imaging 44(2):749–760
9. Yang Z, Simon R, Linte CA (2023) Learning feature descriptors for pre-and intra-operative point cloud matching for laparoscopic liver registration. Int J Comput Assist Radiol Surg 18(6):1025–1032
10. Koo B, Robu MR, Allam M, Pfeiffer M, Thompson S, Gurusamy K, Davidson B, Speidel S, Hawkes D, Stoyanov D et al (2022) Automatic, global registration in laparoscopic liver surgery. Int J Comput Assist Radiol Surg 17(1):167–176
11. Montaña-Brown N, Ramalhinho J, Koo B, Allam M, Davidson B, Gurusamy K, Hu Y, Clarkson MJ (2022) Towards multi-modal self-supervised video and ultrasound pose estimation for laparoscopic liver surgery. In: International workshop on advances in simplifying medical ultrasound, Springer, pp 183–192
12. Zhang H, Bulathsinhala S, Davidson BR, Clarkson MJ, Ramalhinho J (2025) Deep hashing for global registration of preoperative CT and video images for laparoscopic liver surgery. Int J Comput Assist Radiol Surg 20(7):1461–1469
13. Hao J, He B, Dai Y, Li Y, Wang Y, Zhao R, Lian R, Zeng X, Tao H, Yang J et al (2025) A 3d–2d rigid liver registration method using pre-training and transfer learning with staged alignment of anatomical landmarks. Int J Imaging Syst Technol 35(4):70124
14. Gadoux E, Bartoli A (2025) Automatic deep deformable registration using domain adaptation and run-time optimisation. In: International conference on medical image computing and computer-assisted intervention, Springer, pp 65–74
15. Labrunie M, Pizarro D, Tilmant C, Bartoli A (2023) Automatic 3d/2d deformable registration in minimally invasive liver resection using a mesh recovery network. In: MIDL, pp 1104–1123
16. Labrunie M, Ribeiro M, Mourthadhoi F, Tilmant C, Le Roy B, Buc E, Bartoli A (2022) Automatic preoperative 3d model registration in laparoscopic liver resection. Int J Comput Assist Radiol Surg 17(8):1429–1436
17. Zhang H, He L, He R, Kadkhodamohammadi A, Stoyanov D, Davidson BR, Mazomenos EB, Clarkson MJ (2026) Foundationpose-initialized 3d–2d liver registration for surgical augmented reality. https://doi.org/10.48550/arXiv.2602.17517

18. Seo J, Yoo S, Chang J, An H, Ryu H, Lee S, Kruthiventy A, Choi J, Horowitz R (2025) Se (3)-equivariant robot learning and control: a tutorial survey. Int J Control Autom Syst 23(5):1271–1306
19. Barfoot TD (2017) Matrix lie groups. State estimation for robotics. Cambridge University Press, Cambridge, pp 205–284
20. Rabbani N, Calvet L, Espinel Y, Le Roy B, Ribeiro M, Buc E, Bartoli A (2022) A methodology and clinical dataset with ground-truth to evaluate registration accuracy quantitatively in computer-assisted laparoscopic liver resection. Comput Methods Biomech Biomed Eng Imaging Vis 10(4):441–450
21. Mhiri I, Pizarro D, Bartoli A (2025) Neural patient-specific 3d–2d registration in laparoscopic liver resection. Int J Comput Assist Radiol Surg 20(1):57–64